\documentclass{article}

\usepackage[preprint]{neurips_2025}

\usepackage[utf8]{inputenc}
\usepackage[T1]{fontenc}
\usepackage{hyperref}
\usepackage{url}
\usepackage{booktabs}
\usepackage{amsfonts}
\usepackage{nicefrac}
\usepackage{microtype}
\usepackage{xcolor}

\usepackage{graphicx}
\usepackage{pgfplots}
\usepackage{enumitem}
\usepackage{multirow}

\pgfplotsset{width=7cm,compat=1.15}

\usepackage{tipa}

\title{\textsc{iolbench}: Benchmarking LLMs on Linguistic Reasoning}

\author{%
  Satyam Goyal \\
  University of Michigan \\
  \texttt{sagoyal@umich.edu} \\
  \And
  Soham Dan \\
  Microsoft \\
  \texttt{sohamdan@microsoft.com} \\
}

\begin{document}

\maketitle
\begin{abstract}
Despite the remarkable advancements of deep neural networks, their ability to perform reasoning tasks remains limited, particularly in domains requiring structured, abstract thought. This paper addresses a critical gap in current evaluation: the conflation of recalled knowledge with genuine algorithmic reasoning. We introduce \textsc{iolbench}, a benchmark derived from International Linguistics Olympiad (IOL) problems, which uses carefully designed, self-contained linguistic puzzles to isolate core reasoning abilities. By removing the need for external knowledge, these tasks directly evaluate a model's capacity for explicit rule induction and abstract system modeling from sparse data.

Through extensive benchmarking of leading LLMs, we find that even the most advanced models struggle significantly with these tasks, revealing critical weaknesses in compositional generalization and rule abstraction. Our analysis shows these limitations are not random but indicate a fundamental difficulty in abstracting and systematically applying inferred principles. By providing a clear signal on these foundational skills, \textsc{iolbench} serves as a crucial tool for measuring progress towards more robust AI systems, with direct implications for building agents that can reason and operate effectively in novel environments.

\end{abstract}

\begin{figure}[h!]
    \centering
    \includegraphics[width=0.487\textwidth]{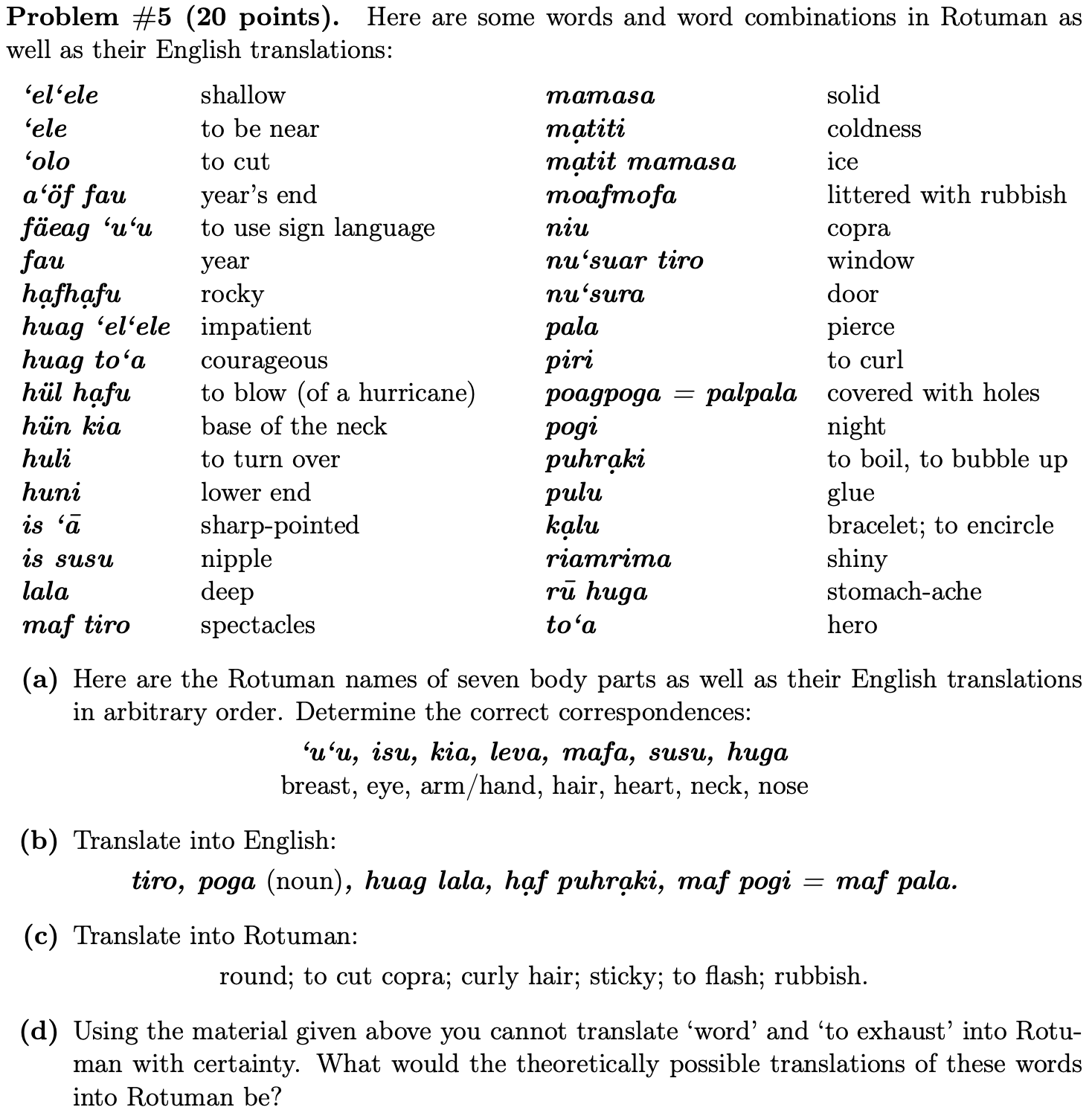}
    \vspace{0.5cm}
    \includegraphics[width=0.487\textwidth]{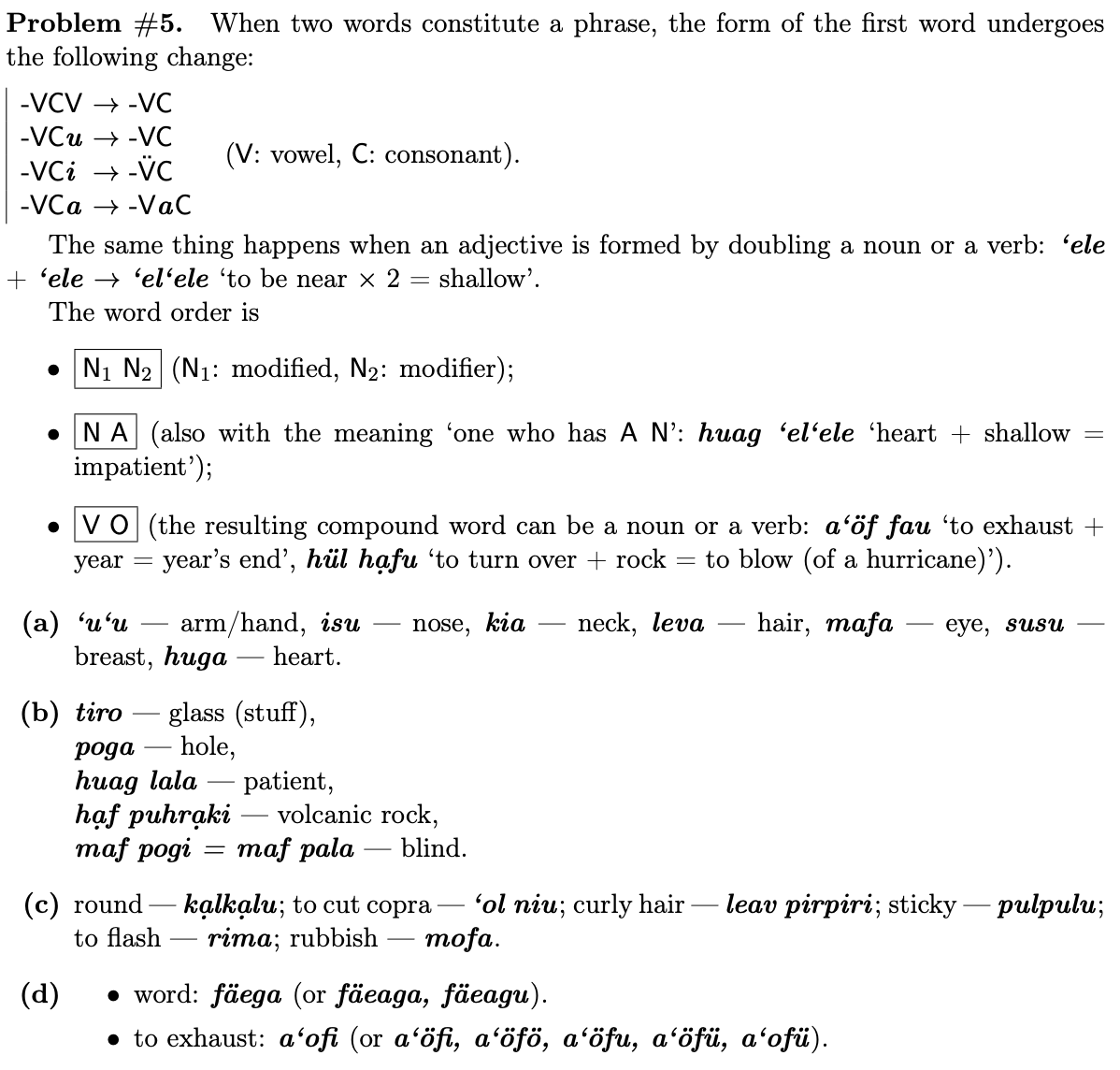}
    \caption{An example problem from IOL 2012 involving the Austronesian language Rotuman, spoken by roughly 9000 people in Fiji. The problem involves lexical matching and translations, and the solution explains how these can be deduced from the provided dictionary and common-sense linguistic reasoning.}
    \label{fig:two-images}
\end{figure}

\section{Introduction}
The emergence of Large Language Models (LLMs) has spurred intense interest in their capacity for reasoning beyond statistical pattern matching. A critical frontier in this research is evaluating a model's ability for genuine algorithmic reasoning, yet a fundamental challenge persists: disentangling this skill from the vast knowledge recalled from training data. Many existing benchmarks, while valuable, inadvertently conflate these two abilities. To truly assess reasoning, we need tasks that neutralize prior knowledge and force models to deduce underlying principles from first principles.

In this work, we introduce \textsc{iolbench}, a benchmark designed to isolate and evaluate these core reasoning faculties. Derived from the International Linguistics Olympiad (IOL), \textsc{iolbench} leverages self-contained, knowledge-independent linguistic puzzles that act as a "clean room" for reasoning evaluation. Specifically, its design allows us to probe three foundational dimensions of reasoning often entangled in other benchmarks: \textbf{explicit rule induction} from sparse data, \textbf{abstract system modeling} of complex interacting parts, and generalization under \textbf{naturalistic complexity}. By using under-documented and typologically diverse languages (e.g., Rotuman in Figure \ref{fig:two-images}), \textsc{iolbench} provides a stringent and focused evaluation of a model's deductive machinery.

To evaluate state-of-the-art performance, we benchmark a suite of leading LLMs, including the GPT-4 family, the Claude 4 family, Gemini 2.5 family, and Grok 4. Our evaluation is guided by two key research questions aligned with our claims: (1) How proficient are leading LLMs at explicit rule induction and abstract system modeling when prior knowledge is neutralized? (2) What systematic failure modes emerge when models are confronted with the naturalistic complexity of diverse linguistic systems? Our experiments analyze performance across task types, including the text-only and multimodal splits, to identify the precise breaking points in model reasoning.

By introducing \textsc{iolbench}, this work offers a targeted diagnostic tool for the AI community. We provide not just a measure of performance, but a clearer signal on the foundational reasoning capabilities and limitations of current models, establishing a crucial foundation for developing robust and generalizable AI systems.

\section{Related Work} Recent efforts to evaluate LLM reasoning have produced a valuable suite of benchmarks. Broad, general-purpose evaluations such as MMLU \citep{hendrycks2020measuring} and BIG-Bench \citep{srivastava2022beyond, suzgun2023challenging} test a wide range of topics from science to social studies. While critical for assessing the breadth of a model's capabilities, these benchmarks inherently struggle with the problem of \textbf{knowledge contamination}. Because their tasks are grounded in real-world concepts potentially seen during pre-training, it is difficult to disentangle genuine, first-principles reasoning from the sophisticated retrieval of learned information. Consequently, they do not provide a clean signal on a model's ability to reason in a true zero-knowledge scenario, a core requirement for operating in novel domains. 

A second category of benchmarks isolates reasoning by focusing on formal, symbolic systems. Datasets for mathematical problem-solving like GSM8K \citep{cobbe2021training}, scientific reasoning like SciBench \citep{wangscibench}, and algorithmic thinking like CLRS \citep{velivckovic2022clrs} evaluate a model's ability to execute structured, multi-step logical procedures. These are invaluable for measuring procedural and logical fidelity. However, their primary focus is on rule \textit{application} within a known, formal system (e.g., the axioms of mathematics or the syntax of a programming language). They do not test the antecedent and arguably more fundamental cognitive skill of \textbf{explicit rule induction}---the ability to \textit{discover} the rules of an unknown system from sparse, ambiguous examples. 

The benchmarks most closely aligned with rule induction are those testing compositional generalization, such as COGS \citep{kim2020cogs}. These datasets effectively test a model's ability to learn a grammar from examples and generalize it to novel combinations. However, they typically rely on synthetic grammars with clean, limited rule sets. This synthetic simplicity contrasts sharply with the layered and often exception-riddled nature of real-world languages. By using problems derived from human languages, \textsc{iolbench} goes a step further, requiring models to generalize not just compositionally, but under the duress of naturalistic complexity. It thereby provides a more rigorous and realistic testbed to model complex, evolving systems.
\section{Benchmark Design and Methodology}

\subsection{The \textsc{iolbench} Dataset}
\textsc{iolbench} is curated from the International Linguistics Olympiad (IOL), a global competition designed to test abstract problem-solving skills rather than prior linguistic knowledge. The design of IOL problems aligns directly with the core reasoning capabilities this paper seeks to evaluate. First, by focusing on low-resource and typologically diverse languages, the problems are inherently self-contained, thus neutralizing the confound of knowledge contamination and ensuring a clean evaluation of first-principles reasoning. Second, each problem requires participants to deduce underlying grammatical principles from a minimal set of examples, directly testing explicit rule induction. Finally, these principles rarely exist in isolation; they form complex, interacting systems with exceptions and hierarchies, providing a rigorous test of generalization under naturalistic complexity.

\subsubsection{Dataset Construction and Reproducibility}
To construct \textsc{iolbench}, we reviewed the complete IOL archive from 2004 to 2023, digitizing approximately 1,500 problem instances from PDFs into a standardized machine-readable format. This meticulous process, which includes transcription, script normalization, and preservation of all associated materials, is crucial for ensuring the benchmark's reproducibility and is detailed further in the GitHub repository. When problems included visual components, these were preserved, creating a multimodal split. Each problem is paired with its official expert solution, enabling fine-grained evaluation of a model's reasoning process. The zero-shot prompt templates used in our experiments are also available in the Appendix.

We partition \textsc{iolbench} into 1198 text-only problems and 52 multimodal problems that require consuming or generating visual information.

\subsubsection{Evaluation Metrics}
Each problem is categorized into one of three types for evaluation:
\begin{enumerate}
    \item \textbf{Type 1 (Short Answer, 666 problems):} Requires a single word or phrase, evaluated via string matching against the gold-standard answer.
    \item \textbf{Type 2 (Long Answer, 501 problems):} Involves longer textual outputs (e.g., translations), evaluated using \textsc{BLEU} and an LLM-based judge on a 0-2 scale.
    \item \textbf{Type 3 (Explanatory Reasoning, 31 problems):} Requires complex explanatory reasoning, graded manually by an expert evaluator on the same 0-2 scale. To ensure scoring consistency, a subset of these problems was graded by a second expert, achieving a high inter-annotator agreement (Cohen's Kappa $\kappa = 0.758$).
\end{enumerate}

\subsection{A Framework for Evaluating Abstract Reasoning}
Beyond model accuracy scores, we analyze model failures through the lens of our three core reasoning dimensions. Below, we define each dimension, establish how we derived it, and provide a canonical example from \textsc{iolbench} that illustrates the required skill.

\subsubsection{Explicit Rule Induction from Sparse Data}
This foundational skill is the ability to observe a pattern in minimal examples and form a hypothesis for a local rule. The capacity for inductive reasoning—generalizing from specific instances to broader principles—is a necessity of both human and artificial general intelligence \citep{goodman2016}. Our analysis will examine cases where models fail at this initial stage of discovery.

The Persian problem (IOL 2007, Problem 1) serves as a clear illustration. The solver is given eight example phrases of the language to its translation and must deduce the semantic difference between two words for "under," \textit{zir-e} and \textit{\textipa{pa:in-e}}.
\begin{quote}
\small
\textbf{Data Snippet:} \\
1. \textit{gorbe zir-e raxtex\textipa{a:}b} – The cat under the bed \\
2. \textit{x\textipa{A}ne p\textipa{a:}in-e kuh} – The house under the mountain \\
... \\
\textbf{Question:} Translate \textit{\textipa{tSaha:rpa:ye pa:in-e miz}} (The stool [?] the table)
\end{quote}
Success on this problem is in part based on observing the data and hypothesizing a grammatical rule: \textit{zir-e} denotes direct physical contact, while \textit{\textipa{pa:in-e}} denotes a general lower elevation. This act of hypothesizing one core principle from a few contrasting examples is explicit rule induction.

\subsubsection{Abstract System Modeling}
This skill involves synthesizing multiple, distinct rules into a coherent, interacting system. The ability to form such internal "mental models" has long been considered a central component of complex cognition and problem-solving in humans \citep{johnson-laird1983mental}. A failure here indicates a model can identify individual patterns but cannot grasp how they fit together.

The Ancient Greek problem (IOL 2007, Problem 2) is a perfect test of this skill. A solver must match phrases to translations by building a mini-grammar for noun phrases.
\begin{quote}
\small
\textbf{Data Snippet:} \\
A. \textit{ho t\textipa{o:}n hyi\textipa{o:}n dulos} \\
B. \textit{hoi t\textipa{o:}n dul\textipa{o:}n cyrioi} \\
... \\
\textbf{Translations (Unordered):} \\
1. the slave of the sons \\
2. the masters of the slaves \\
...
\end{quote}
Solving this requires building a multi-part system model by correctly inducing and integrating several interlocking rules: (1) a rule for number distinguishing \textit{ho} (sg.) from \textit{hoi} (pl.); (2) a rule for possession distinguishing \textit{tu} (of the sg.) from \textit{t\textipa{o:}n} (of the pl.); and (3) a rule for word order. A solver cannot succeed by finding one trick; they must model how the number, possession, and word-order subsystems interact to form a coherent whole.

\subsubsection{Generalization under Naturalistic Complexity}
This dimension tests the ability of an inferred system to handle the messy, irregular, and layered nature of real-world phenomena. The algebraic capacity to compose known rules is a key feature of general intelligence \citep{fodor1988connectionism}, however, real-world systems often contain exceptions and interactions not present in clean, formal systems. This skill evaluates the robustness of a model's reasoning when moving beyond idealized scenarios.

The Tupi-Guarani problem (IOL 2011, Problem 3) exemplifies this challenge. It requires inducing a series of sound changes between the related Tupinamb\textipa{\'a} and Guarani Mbya languages.
\begin{table}[h!]
\centering
\small
\begin{tabular}{|l|l|l|}
\hline
\textbf{English} & \textbf{Tupinamb\textipa{\'a}} & \textbf{Guaran\textipa{\'i} Mbya} \\
\hline
rock & it\textipa{\'a} & it\textipa{\'a} \\
head & akanga & ak\textipa{\~a} \\
to bring & erur & eru \\
you want & erepot\textipa{\'a}r & erepot\textipa{\'a} \\
to heal & puer\textipa{\'a}b & kuer\textipa{\'a} \\
\hline
\end{tabular}
\end{table}
The task involves filling in blank words by applying the inferred rules. This requires not just building a system model of sound changes, but generalizing this model to new words that introduce the complexity of real language evolution. The sound changes are not perfectly regular; they interact with syllable structure and vowel quality. This forces the solver to apply their inferred system to the noisy, irregular data characteristic of a natural, evolved system.

\section{Experiments and Results}

\subsection{Experimental Setup}
\textbf{Models Evaluated.} We evaluate a suite of state-of-the-art LLMs to assess the current frontier of linguistic reasoning. The models include: OpenAI's GPT-4 family (GPT-4, GPT-4o, GPT-4o mini), OpenAI's o1 model, Anthropic's Claude 4 family (Sonnet and Opus), xAI's Grok-4, and Google's Gemini 2.5 family (Pro and Flash).

\textbf{Prompting Strategy.} All experiments are conducted using a zero-shot prompting strategy. Each problem from \textsc{iolbench} is presented to the model with instructions to solve it based solely on the provided information. This approach is critical for enforcing the knowledge-independent nature of our evaluation. The complete prompt templates used for each problem type are available in Appendix.

\subsection{Quantitative Analysis}
The overall performance of the evaluated models on the text-only and multimodal splits of \textsc{iolbench} is summarized in Table \ref{tab:combined_performance_metrics} and Table \ref{tab:combined_vision_metrics}, respectively. These results provide a high-level overview of current capabilities and highlight the significant challenge our benchmark presents.

\begin{table}[h!]
\centering
\small
\begin{tabular}{|l||c||c|c||c|c|}
\hline
\multirow{2}{*}{\textbf{Model}} & 
\textbf{Type 1} & \multicolumn{2}{c||}{\textbf{Type 2}} & \multicolumn{1}{c|}{\textbf{Type 3}} \\
\cline{2-5}
& \textbf{Acc.} & \textbf{LLM} & \textbf{Bleu} & \textbf{Manual }\\
\hline\hline
GPT-5          & \textbf{61.12} & 41.91 & 20.76 &  20.21\\ \hline
Grok-4         & 38.11 & 32.91 & \textbf{22.86} &  25.21\\ \hline
Claude 4 sonnet      & 36.84 & 46.07 & 27.18 & 16.17 \\ \hline
Claude 4 opus     & 41.90 & 46.45 & 20.91 & 28.88 \\ \hline
GPT-4          & 24.48 & 37.70 & 10.81 & 20.25 \\ \hline
GPT-4o         & 26.87 & 39.25 & 7.30  & 27.29 \\ \hline
GPT-4o mini      & 20.62 & 31.17 & 6.20  & 19.83 \\ \hline
Gemini 2.5 pro & 39.76 & \textbf{50.47} & 37.05  & \textbf{29.79} \\ \hline
Gemini 2.5 flash   & 19.76 & 35.47 & 17.75  & 19.79 \\ \hline
\end{tabular}

\caption{Performances (in \%) of different LLMs for each problem category for text-problems in \textsc{iolbench}. The best results for each category are bolded.}
\label{tab:combined_performance_metrics}
\end{table}

\begin{table}[h!]
\centering
\small
\begin{tabular}{|l||c||c|}
\hline
\multirow{1}{*}{\textbf{Model}} & 
\textbf{Type 1} & \textbf{Type 2} \\
\cline{2-3}
\hline\hline
Claude 4 Sonnet & 33.10 & 24.20 \\ \hline
Grok-4 & 27.10 & 12.22 \\ \hline
Claude 4 Opus & 44.80 & \textbf{43.10} \\ \hline
GPT-4o & 10.00 & 17.50 \\ \hline
Gemini 2.5 Pro & \textbf{45.20} & 39.10 \\ \hline
\end{tabular}
\caption{Performances (in \%) of different LLMs for each problem category for multimodal-problems in \textsc{iolbench}.}
\label{tab:combined_vision_metrics}
\end{table}

As the tables indicate, overall performance remains low across all models, confirming that the abstract reasoning required by \textsc{iolbench} is a significant challenge for current architectures. Gemini 2.5 Pro demonstrates the strongest performance on text-based tasks, while Claude 4 Opus excels on the more difficult multimodal problems. However, to understand the nature of these limitations, we now turn to a qualitative analysis of model failures.

\subsection{Qualitative Error Analysis: A Dimensional Breakdown}
\label{sec:qualitative}
Quantitative scores reveal that models struggle; a qualitative analysis is necessary to understand the dimensionality of failures. We now analyze specific model failures through the lens of our three reasoning dimensions, using representative examples to diagnose the breaking points in the models' reasoning processes.

\subsubsection{Failure in Explicit Rule Induction}
The most fundamental errors occur when models fail to identify simple, local patterns, demonstrating an inability to perform the discovery step of reasoning. In these cases, models often hallucinate non-existent rules or overlook clear evidence in the data. The Lalana Chinantec problem (IOL 2008, Problem 5) provides an example of this failure. The task requires decomposing sentences to find the meaning of individual morphemes.

\begin{figure}[h!]
\centering
\fbox{
\begin{minipage}{0.9\linewidth}
\small
\textbf{Problem Snippet (Lalana Chinantec):} \\
\texttt{1. kalakwa: kw\textipa{1}: li:\textipa{P} – The beautiful corn grew.} \\
\texttt{2. m\textipa{1}la\textipa{dZ}\"{o} mo:h kya – My pineapples have turned out well.} \\
\texttt{3. li:\textipa{P} kalane kw\textipa{1}: kwa: kya – My tall corn yellowed beautifully.} \\
... \\
\textbf{Question:} What does the word \texttt{li:\textipa{P}} mean?

\vspace{5pt}
\hrule
\vspace{5pt}

\textbf{Model Response (Gemini 2.5 Pro (thinking)):} \\
\texttt{...This suggests li:\textipa{P} can mean both "grew" and "yellowed" - likely it means "yellowed" and in sentence 1 it's part of a compound meaning.} \\
\texttt{Final Analysis: Part a: li:\textipa{P} means "yellowed"}
\end{minipage}
}
\caption{A failure of explicit rule induction. The model gets confused by the polysemy in the English translation and fails to deduce the correct meaning of \texttt{li:\textipa{P}} from the syntactic context.}
\label{fig:rule_induction_failure}
\end{figure}

The model's reasoning in Figure \ref{fig:rule_induction_failure} demonstrates a failure of rule induction. By comparing sentences 1 and 3, a human solver can deduce that verbs can appear before or after the noun phrase. In sentence 1, \texttt{li:\textipa{P}} is the verb `grew`, while in sentence 3, the verb is `kwa:` (`yellowed`). The model, however, incorrectly assumes a fixed word order and gets confused, ultimately misidentifying \texttt{li:\textipa{P}} as `yellowed`. It failed to ground its reasoning in the syntactic evidence provided, instead using a plausible but incorrect semantic association.

\subsubsection{Failure in Abstract System Modeling}
A more complex failure occurs when a model correctly identifies some individual rules but cannot synthesize them into a coherent, interacting system. The Guaran\textipa{\'i} Verbs problem (IOL 2004, Problem 16) highlights this weakness. Solving it requires building a complete model of the verb paradigm from multiple, interlocking morphemes.

\begin{figure}[h!]
\centering
\fbox{
\begin{minipage}{0.9\linewidth}
\small
\textbf{Problem Snippet (Guaran\textipa{\'i} Verbs):} \\
\texttt{japurahei – We are singing} \\
\texttt{no\textipa{\~n}e'\textipa{\~e}i – He is not talking} \\
\texttt{ndajajupirima – We were not waking up} \\
... \\
\textbf{Question:} Translate "We will be eating" into Guaran\textipa{\'i}.

\vspace{5pt}
\hrule
\vspace{5pt}

\textbf{Model Response (Claude 4 Opus (Thinking)):} \\
\texttt{...Morphological Analysis: Looking at negation patterns... The future appears to use mo'\textipa{\~a} as an auxiliary or suffix.} \\
\texttt{Part b - Guaran\textipa{\'i} translations: ... "We will be eating" = jakarumo'\textipa{\~a}ta (ja-karu-mo'\textipa{\~a}-ta)}
\end{minipage}
}
\caption{A failure of abstract system modeling. The model correctly identifies individual morphemes but fails to synthesize the correct interaction for future tense, incorrectly combining a future auxiliary with a future suffix.}
\label{fig:system_modeling_failure}
\end{figure}

As shown in Figure \ref{fig:system_modeling_failure}, the model successfully performs rule induction on the person prefixes (\textit{ja-} for "we") and the verb stem (\textit{karu} for "eat"). However, it fails to build a correct system model for tense. It observes both a future suffix \textit{-ta} and a future auxiliary \textit{mo'\textipa{\~a}} (used in negative constructions) and incorrectly concludes that they should be combined. The correct answer, \textit{jakaruta}, requires understanding that these are part of separate subsystems (affirmative vs. negative future tense). The model identified the parts but could not assemble them correctly, demonstrating a critical failure in synthesizing rules into an accurate system.

\subsubsection{Failure of Generalization under Naturalistic Complexity}
The most advanced reasoning failures occur when a model's inferred system, while perhaps internally consistent, doesn't generalize well to complex linguistic structure. The Icelandic Kinship problem (IOL 2005, Problem 14) demonstrates this, requiring a model to apply the rules of Icelandic naming to a multi-generational family tree.

\begin{figure}[h!]
\centering
\fbox{
\begin{minipage}{0.9\linewidth}
\small
\textbf{Problem Snippet (Icelandic Kinship):} \\
\texttt{...Gu\textipa{D}r\textipa{\'u}n Eriksd\textipa{\'o}ttir and Jakob Christiansson had three kids... Here is the list of all the people in the party: Christian Krist\textipa{\'i}nsson, Daniel Gu\textipa{D}r\textipa{\'u}narson Jakobsson, ... Krist\textipa{\'i}n Gu\textipa{D}r\textipa{\'u}nard\textipa{\'o}ttir Aronsd\textipa{\'o}ttir...} \\
\textbf{Question:} Which of the sons/daughters of Gu\textipa{D}r\textipa{\'u}n and Jakob has no grandsons/granddaughters?

\vspace{5pt}
\hrule
\vspace{5pt}

\textbf{Model Response (Grok-4 (thinking)):} \\
\texttt{"2": "Krist\textipa{\'i}n Gu\textipa{D}r\textipa{\'u}nard\textipa{\'o}ttir Aronsd\textipa{\'o}ttir"}
\end{minipage}
}
\caption{A failure to generalize to naturalistic complexity. The model incorrectly identifies Krist\textipa{\'i}n as a child of Gu\textipa{D}r\textipa{\'u}n, failing to parse the multi-part surnames.}
\label{fig:naturalistic_failure}
\end{figure}

In Figure \ref{fig:naturalistic_failure}, the model fails a complex inferential task. To answer correctly, one must first identify all of Gu\textipa{D}r\textipa{\'u}n and Jakob's children from the list, then identify all of their children (the grandchildren), and finally see which of the original children has no descendants in the third generation. The correct answer is Ragnhei\textipa{D}ur Jakobsd\textipa{\'o}ttir. The model incorrectly identifies Krist\textipa{\'i}n Gu\textipa{D}r\textipa{\'u}nard\textipa{\'o}ttir Aronsd\textipa{\'o}ttir. This error reveals that the model's internal system for Icelandic names was not generalizable. It likely identified "Gu\textipa{D}r\textipa{\'u}nard\textipa{\'o}ttir" (daughter of Gu\textipa{D}r\textipa{\'u}n) however Krist\textipa{\'i}n also has a second matronymic, "Aronsd\textipa{\'o}ttir," indicating her mother is someone named Aron, not Gu\textipa{D}r\textipa{\'u}n. The model's derived rules fail to generalize under naturalistic complexity.

\section{Discussion and Conclusion}
In this work, we introduced \textsc{iolbench}, a benchmark designed not just to measure performance on linguistic puzzles, but to isolate and diagnose fundamental reasoning capabilities in Large Language Models. Our experiments reveal that even state-of-the-art models exhibit significant weaknesses when confronted with tasks requiring genuine, knowledge-independent reasoning. The quantitative results show low aggregate scores, but our qualitative analysis provides the critical insight on three key dimensions of reasoning where LLMs fail. We find that models struggle with the foundational skill of explicit rule induction, often fail to synthesize discovered rules into a coherent abstract system model, and struggle to form generalizable reasoning strategies against naturalistic complexity.

These findings have profound implications for the development of advanced AI. The goal of creating autonomous agents that can reason, plan, and act in novel environments is a central theme in current research \citep{bubeck2023sparks}. Such an agent's success is predicated on its ability to do precisely what IOLBENCH tests: deduce the "rules of the world" from sparse observations. Our results suggest that current architectures, despite their vast knowledge, lack the robust deductive machinery to do this reliably. A model that cannot induce the simple rules of Rotuman morphology is unlikely to reliably infer the complex causal rules of a new scientific domain or an unfamiliar interactive environment. The failures we document are therefore not just linguistic, but are indicative of a deeper architectural gap that must be addressed to build more capable and trustworthy AI systems.

\textbf{Limitations and Future Work.} Our analysis provides a clear diagnostic, but it is not exhaustive. The current work focuses on a representative set of leading proprietary models; future work should expand this evaluation to a wider range of open-source models and varied architectures to broaden our understanding. Furthermore, while we focused on zero-shot evaluation to ensure a clean test of reasoning, exploring the efficacy of few-shot prompting and fine-tuning on a subset of IOLBENCH could reveal important insights into how these reasoning skills might be learned. Additionally, expanding the \textsc{iolbench} dataset across more questions, especially multimodal ones, can yield more insights on model behavior. We release \textsc{iolbench} and its code to the community to facilitate this research, providing a challenging and precise tool to measure our collective progress towards true artificial general intelligence.

\bibliography{custom}
\bibliographystyle{plainnat}

\appendix
\section{Prompt Template for Zero-Shot Evaluation}
\label{sec:appendix}

All experiments were conducted using a zero-shot prompting strategy. The models were given a persona and a structured set of instructions to guide their reasoning and formatting. The complete prompt template used for querying the large language models is provided below. The specific problem text was inserted into the $`<problem>`$ tag, and the required JSON structure was provided in the tag $`<answer_format>`$ for each problem.

\vspace{10pt}

\begin{center}
\fbox{
\begin{minipage}{0.9\linewidth}
\small
\texttt{As a linguistics expert with deep knowledge of computational linguistics, historical linguistics, and language structure analysis, solve the following problem systematically.}

\texttt{Analyze the linguistic patterns, apply your knowledge of phonology, morphology, syntax, and semantics as appropriate. Show your reasoning clearly and provide accurate solutions.}

\texttt{At the end, provide your solution in the exact JSON format specified below. Include only the answers in the final JSON - no explanations or additional text after the JSON format.}

\texttt{<answer\_format>} \\
\texttt{\{} \\
\texttt{  "Problem X": \{} \\
\texttt{    "Part a": \{} \\
\texttt{      ...} \\
\texttt{    \}} \\
\texttt{  \}} \\
\texttt{\}} \\
\texttt{</answer\_format>}

\texttt{<problem>} \\
\texttt{[Problem text is inserted here]} \\
\texttt{</problem>}
\end{minipage}
}
\end{center}

\end{document}